%% file: main.tex
\documentclass{article}

\usepackage[english]{babel}
\usepackage{amsfonts}
\usepackage{mathbbol}
\usepackage[letterpaper,top=2cm,bottom=2cm,left=3cm,right=3cm,marginparwidth=1.75cm]{geometry}

\PassOptionsToPackage{numbers, compress}{natbib}
\usepackage[]{natbib}

\newcommand\blfootnote[1]{%
  \begingroup
  \renewcommand\thefootnote{}\footnote{#1}%
  \addtocounter{footnote}{-1}%
  \endgroup
}
\usepackage[letterpaper,top=2cm,bottom=2cm,left=3cm,right=3cm,marginparwidth=1.75cm]{geometry}

\usepackage[colorlinks=true, allcolors=blue]{hyperref}
\DeclareMathAlphabet      {\mathbfit}{OML}{cmm}{b}{it}

\usepackage{amsmath}
\usepackage{graphicx}
\usepackage[colorlinks=true, allcolors=blue]{hyperref}

\title{Unfamiliar Finetuning Examples Control How \\ Language Models Hallucinate}
\author{Katie Kang$^1$, Eric Wallace$^1$, Claire Tomlin$^1$, Aviral Kumar$^2$, Sergey Levine$^1$}
\date{%
    $^1$UC Berkeley   
    $^2$Google DeepMind%
}
\begin{document}
\maketitle

\begin{abstract}
Large language models are known to hallucinate when faced with unfamiliar queries, but the underlying mechanism that govern how models hallucinate are not yet fully understood. In this work, we find that unfamiliar examples in the models' finetuning data -- those that introduce concepts beyond the base model's scope of knowledge -- are crucial in shaping these errors. In particular, we find that an LLM's hallucinated predictions tend to mirror the responses associated with its unfamiliar finetuning examples. This suggests that by modifying how unfamiliar finetuning examples are supervised, we can influence a model's responses to unfamiliar queries (e.g., say ``I don't know''). We empirically validate this observation in a series of controlled experiments involving SFT, RL, and reward model finetuning on TriviaQA and MMLU. Our work further investigates RL finetuning strategies for improving the factuality of long-form model generations. We find that, while hallucinations from the reward model can significantly undermine the effectiveness of RL factuality finetuning, strategically controlling how reward models hallucinate can minimize these negative effects. Leveraging our previous observations on controlling hallucinations, we propose an approach for learning more reliable reward models, and show that they improve the efficacy of RL factuality finetuning in long-form biography and book/movie plot generation tasks. \blfootnote{Correspondence to: \href{mailto:katiekang@eecs.berkeley.edu}{katiekang@eecs.berkeley.edu}}\blfootnote{
Code: \url{https://github.com/katiekang1998/llm_hallucinations}}
\end{abstract}

\section{Introduction}

Large language models (LLMs) have a tendency to ``hallucinate,'' generating plausible-sounding responses that are factually incorrect. This behavior is especially prominent when models are queried on concepts that extend beyond the models' knowledge base~\cite{kandpal2023large, kalai2023calibrated} (e.g., asking the model to generate the biography of a little-known person). We will refer to these queries as \emph{unfamiliar} inputs. Rather than fabricating information when presented with unfamiliar inputs, models should instead 
verbalize their uncertainty or confine their responses within the limits of their knowledge. The goal of our work is to teach models this behavior, particularly for long-form generation tasks. 

\begin{figure}
  \centering
  \includegraphics[width=\textwidth]{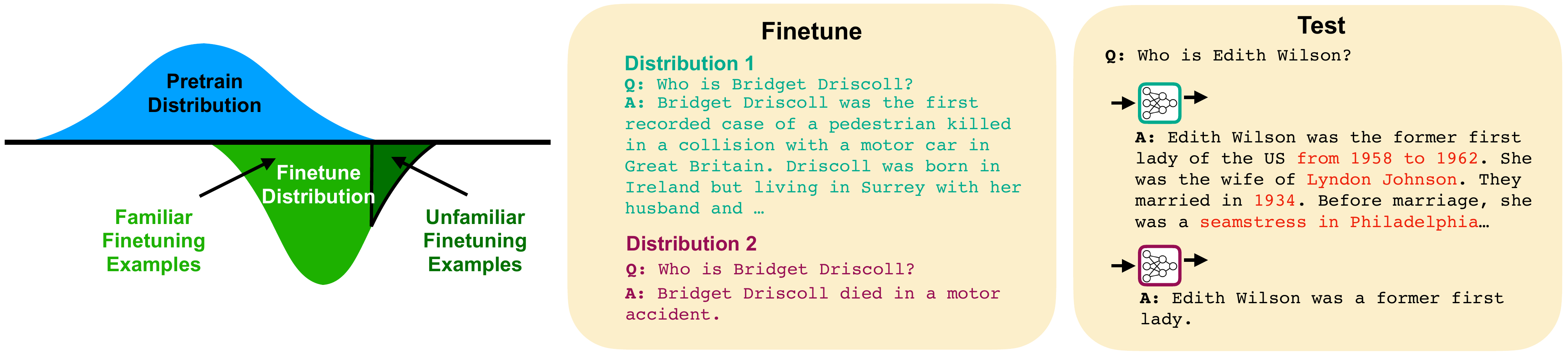}
  \caption{Conceptual visualization of (un)familiar finetuning examples (left), and example of model predictions mimicking unfamiliar finetuning examples (middle and right). When finetuning on distribution 1, which contains details the model may not know, the model outputs detailed responses at test-time with inaccuracies (red). When finetuning on distribution 2, which omits unfamiliar details, the model produces shorter responses with fewer inaccuracies. }
  \label{fig:teaser}
  \vspace{-13pt}
\end{figure}

Towards this goal, we first set out to better understand the underlying mechanisms that govern how LLMs hallucinate. Our investigation reveals that a finetuned model's hallucinated responses tend to mimic the unfamiliar examples the model's finetuning data (i.e., finetuning examples containing concepts unfamiliar to the pretrained model). More specifically, as test queries become more unfamiliar, we find that LLM predictions tend to default toward the distribution of responses associated with the model's unfamiliar finetuning examples. We illustrate this observation with an example in Fig. \ref{fig:teaser}. To empirically verify this phenomenon, we conduct a series of controlled experiments, where we manipulate the way unfamiliar finetuning examples are supervised, and investigate the effect on the finetuned model's predictions. We use multiple-choice (MMLU) and short-form question answering tasks (TriviaQA) as testbeds, where we can precisely characterize an LLM's output distribution. Our results show that, across different finetuning procedures including SFT, RL, and reward model finetuning, the model predictions for unfamiliar test queries indeed approach the distribution of responses in the model's unfamiliar finetuning examples.

Our observation suggests a recipe for minimizing factual inaccuracies in model generations: by strategically manipulating the unfamiliar examples in the model's finetuning data, we can steer the model's predictions for unfamiliar queries towards more desirable (e.g. linguistically uncertain) responses. We leverage this insight to design better finetuning techniques to improve the factuality of long-form LLM generations. In particular, our study focuses on RL-based approaches, where the use of reward models to supervise finetuning makes it scalable to long-form tasks. However, reward models themselves can suffer from hallucinations in the face of unfamiliar inputs, which can diminish the efficacy of RL factuality finetuning. To tackle this challenge, we draw on our previous insights to strategically control how reward models hallucinate. In particular, we find that overestimated reward predictions tend to be more harmful than underestimated reward predictions, and propose an approach for learning reward models that avoid overestimating rewards for unfamiliar inputs, which we call conservative reward models. On biography and book/movie plot generation tasks, we find that using conservative reward models for RL factuality finetuning can significantly reduce the adverse effects of reward hallucinations, and that this approach can more reliably teach models to generate factual long-form responses than standard SFT and RL with standard reward models. 

In summary, our work makes two primary contributions: (1) we present a conceptual model outlining the factors that influence finetuned LLM predictions in response to unfamiliar queries, and (2) we leverage our findings to develop a more reliable approach to RL factuality finetuning for long-form generation tasks. We hope that the insights in our paper contribute to a better understanding of the mechanisms that govern how LLMs hallucinate, and the principles for controlling these hallucinations.

\section{Related Work}
A number of works have documented the tendency of LLMs to hallucinate factually incorrect responses~\cite{kalai2023calibrated, bubeck2023sparks, kadavath2022language, agrawal2023language}. Additionally, studies have investigated the conditions under which hallucinations occur and how LLMs behave in such instances. In particular, LLMs tend to hallucinate more frequently when queried on knowledge that is rarely mentioned in their training data~\cite{mallen2023not, kandpal2023large}. Furthermore, LLM predictions generally tend to be moderately calibrated~\cite{kadavath2022language, zhao2021calibrate, tian2023just}, and their internal representations seem to reflect some awareness of model uncertainty~\cite{liu2023cognitive, azaria2023internal}. Our work, which finds that LLM hallucinations mimic the responses associated with its unfamiliar finetuning examples, extends our understanding of LLM behavior under uncertainty.

Prior work has observed phenomena similar to our observation in standard neural networks (those without pretraining)~\cite{kang2023deep, hendrycks2016baseline}. These works show that, as inputs become more out-of-distribution, neural network predictions tend to default towards a predictable value --- much like the default behavior of LLMs when faced with unfamiliar queries. However, because standard neural networks lack the initial foundation of a pretrained model, the constant prediction reflects the model's training distribution rather than the unfamiliar examples encountered during finetuning.

Finally, a number of prior works have similarly sought to address the challenges posed by LLM hallucinations. Active research areas include hallucination detection~\cite{manakul2023selfcheckgpt, mundler2023self, xu2023understanding, kuhn2023semantic}, automated evaluation of factuality~\cite{min2023factscore, umapathi2023med, jing2023faithscore}, and mitigation techniques. Common strategies for mitigating hallucinations include specialized sampling methods~\cite{lee2022factuality, li2023inference, chuang2023dola, zhang2023alleviating}, more reliable input prompts~\cite{si2022prompting}, using retrieval augmentation to incorporate external knowledge~\cite{gao2023rarr, peng2023check, varshney2023stitch, yaowikichat, shuster2021retrieval}, and, closest to our work, finetuning models for factuality. In particular, prior works has found that SFT on data where difficult examples are labeled to abstaining answers~\cite{lin2022teaching, yang2023alignment, zhang2023r}, as well as RL finetuning~\cite{shulman, goldberg, tian2023fine, sun2023aligning, roit2023factually, mesgar2020improving} can improve the factuality of model generations, which we also observe in our experiments. While these works propose specific approaches for tackling hallucinations, our work instead aims to better understand the underlying mechanisms that govern language models hallucinations in a unified manner. Furthermore, our work investigates the little-studied effects of reward model hallucinations, which we find to have a large impact on the efficacy of RL factuality finetuning. 

\section{Problem Setting}
Modern LLMs are typically trained in a two-stage process: pretraining on broad-coverage corpora, followed by finetuning on more specialized instruction-following datasets~\cite{ouyang2022training}. These models are prone to generating undesirable responses when prompted with inputs that are not well represented in their training data. In particular, models tend to output plausible-sounding but factually incorrect responses when queried outside its pretraining distribution, and output nonsensical responses when queried outside its finetuning distribution. We focus on the former regime of hallucinations, where queries stylistically resemble examples in the finetuning data, but require concepts beyond the pretrained model's scope of knowledge. We call this kind of input \emph{unfamiliar} to the model. 

In our experiments, we will use question-answer tasks as a testbed, though our analysis and method can apply to any prompted generation LLM task. To isolate the effects of distribution shift with respect to the pretraining data (rather than finetuning data), we will evaluate model predictions on held-out queries sampled from the same distribution as the finetuning data. To understand how the behavior of the model changes depending on the unfamiliarity of the test query, our evaluation will decompose the held-out test set into different levels of unfamiliarity. We will quantify the unfamiliarity of a query by few-shot prompting the pretrained model with a few examples (sampled from the same task) along with the query of interest, and measuring the quality of the pretrained model's prediction, where the quality of a prediction is quantified using task-specific metrics. We refer to this metric as the unfamiliarity score of a query. We consider a finetuning example to be unfamiliar if the unfamiliarity score of its query is above a certain threshold, and familiar otherwise. 

 \section{Understanding How LLMs Hallucinate}
\label{sec:understanding}
In this section, we investigate the underlying mechanisms that govern how finetuned LLMs hallucinate. We hypothesize that, when face with unfamiliar inputs, model predictions mimic the responses associated with the model's unfamiliar finetuning examples. We will first present our hypothesis more precisely, then validate our hypothesis with a series of controlled experiments. 

\subsection{Main Hypothesis} 
Let us consider an LLM $f_\theta$, which maps a prompt $x$ to a distribution of responses $P_\theta(y|x)$. We finetune this model on a dataset $\mathcal{D} = \{(x_i, \mathbfit{s}_i)\}_{1 \leq i \leq N}$
with a loss function $\sum_{(x_i, \mathbfit{s}_i) \in \mathcal{D}} \mathcal{L}(f_\theta(x_i),  \mathbfit{s}_i)$, where $s_i$ represents the supervision associated with $x_i$. Depending on the choice of $\mathcal{L}$, this can represent SFT (where $\mathbfit{s}_i$ is a a target response) or RL finetuning (where $\mathbfit{s}_i$ is a reward function). 

While the optimal behavior that an LLM can learn during finetuning is to output the ground-truth answer to each query, this may not happen in practice for all finetuning examples. For familiar finetuning examples, the pretrained model's representations often encode useful associations between queries and responses, facilitating the finetuning optimization for those examples. However, for unfamiliar examples, which we refer to as $\mathcal{D}_{\text{unf}}$, such helpful associations in the pretrained representations are largely absent, making it more difficult to model these examples. Nonetheless, while an LLM may struggle to produce the optimal response for each query in $\mathcal{D}_{\text{unf}}$, it can still reduce the finetuning loss by learning to predict the types of responses associated with unfamiliar examples. More specifically, the model can minimize the \emph{aggregate} loss over unfamiliar finetuning examples by producing an intelligent ``blind guess'', $P_{\text{unf}}(y) = \arg\min_{P(y)} \sum_{(x_i, \mathbfit{s}_i) \in \mathcal{D}_{\text{unf}}} \mathcal{L}(P(y), \mathbfit{s}_i)$, for all unfamiliar queries. Note that $P_{\text{unf}}(y)$ is input-agnostic, and depends only on the model's unfamiliar finetuning examples. We hypothesize that \textbf{LLMs learn to predict this intelligent ``blind guess''  ($P_{\text{unf}}(y)$) for unfamiliar examples during finetuning, and that they default to this prediction when faced with unfamiliar queries at test time}. 

\subsection{Experimental Verification of our Main Hypothesis}
We will now present a series of experiments to evaluate our hypothesis. The goal of our experiments is to verify that {(1)} model predictions indeed default to $P_{\text{unf}}(y)$ when presented with unfamiliar queries, and {(2)} this prediction behavior is controlled by the unfamiliar examples in the models' finetuning data. Towards this goal, we analyze the prediction behavior of different models, where unfamiliar finetuning examples are supervised in different ways, while all other training details are kept fixed. To evaluate our hypothesis for different types of finetuning procedures, we finetune models to generate responses using both SFT and RL, as well as to predict rewards (as reward models for RL finetuning). We use Llama2 7B~\cite{touvron2023llama} as the pretrained model. We conduct our experiments with a multiple-choice (MMLU~\cite{hendrycks2020measuring}) and a short-form (TriviaQA~\cite{joshi2017triviaqa}) question answering task, so that we can precisely characterize a model's output distributions. For MMLU, we obtain the unfamiliarity score by few-shot prompting the pretrained model and measuring the  negative log likelihood of the correct answer under the predicted distribution. For TriviaQA, we obtain the unfamiliarity score by few-shot prompting the pretrained model, sampling 12 responses, and measuring the number of incorrect responses. In subsequent sections, we will extend our experiments to long-form generation tasks. For further experimental details, see Appendix \ref{appdx:mmlu} and \ref{appdx:triviaqa}.


\begin{figure}
  \centering
  \includegraphics[width=\textwidth]{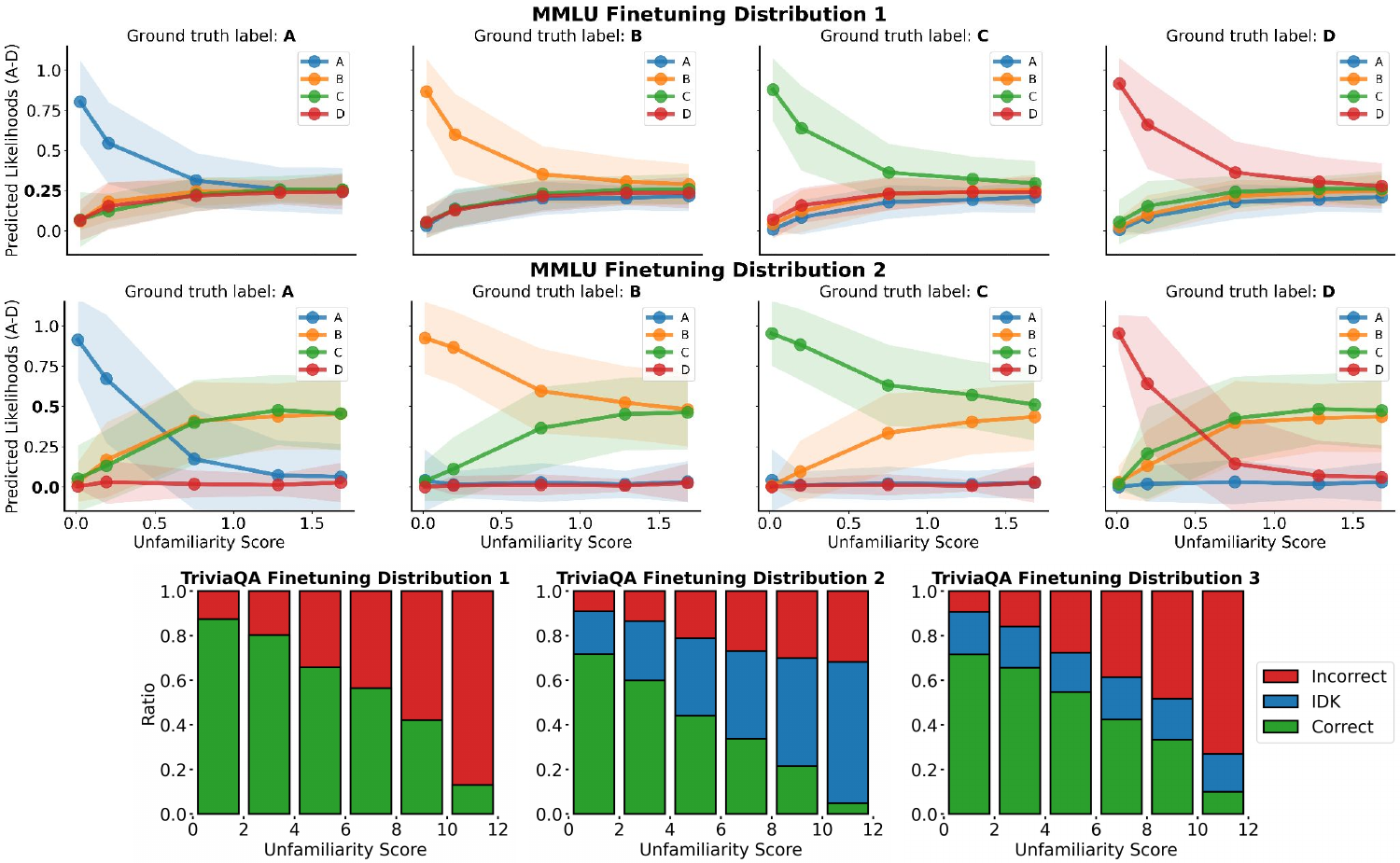}
  \caption{Prediction behavior of models finetuned with SFT on MMLU (top 2 rows) and TriviaQA (bottom row). For MMLU plots, only test inputs with a specific ground truth label (A-D) are evaluated within each column. Solid line represents the average predicted likelihood, and error bars represent standard deviation within the test set. For TriviaQA plots, each bar denotes the ratio of model outputs within each category. For all plots in this figure, as inputs become more unfamiliar, model predictions default towards the distribution of target responses in the model's unfamiliar finetuning examples.}
  \label{fig:sft}
  \vspace{-10pt}
\end{figure}

\textbf{Supervised finetuning.} First, we investigate the prediction behavior of models finetuned with SFT to predict responses to input queries. For this training objective, $P_{\text{unf}}(y)$ corresponds to the marginal distribution of target responses in the set of unfamiliar finetuning examples. 

In our experiments with MMLU, we consider two different finetuning data distributions. In the first distribution, the target responses in both familiar and unfamiliar examples are distributed uniformly over A-D tokens. In the second distribution, the target responses in familiar examples are distributed uniformly, while the target responses in unfamiliar examples are distributed 50\% B and 50\% C. For a model finetuned on the first data distribution,  $P_{\text{unf}}(y)$ corresponds to the uniform distribution over A-D, while for a model finetuned on the second distribution, $P_{\text{unf}}(y)$ corresponds to 50\% B/50\% C. In the top of Fig. \ref{fig:sft}, we plot the two models' predicted distributions over A-D as their test inputs become more unfamiliar (left to right on the x-axis). We can see that for familiar test inputs, both models predicted higher likelihoods for the letter associated with the ground truth answer. However, as inputs become more unfamiliar, the predictions of the first model approached the uniform distribution, while the predictions of the second model approached the 50\% B/50\% C distribution.

In our experiments with TriviaQA, we consider three different finetuning data distributions. In the first, all finetuning examples are labeled with the ground-truth answer to their respective queries. In the second, familiar examples are labeled with the ground-truth answer, while unfamiliar examples are labeled with ``I don't know''. In the third, a random subset of examples are labeled with ``I don't know'' and with rest are labeled with the ground-truth answer, where the ratio of examples with ``I don't know'' labels matches that of the second data distribution. For models finetuned on these distributions, responses from $P_{\text{unf}}(y)$ correspond to hallucinated answers, ``I don't know'', and a mixture of hallucinated answers and ``I don't know'', respectively. In the bottom of Fig. \ref{fig:sft}, we visualize sampled responses from the three models. Comparing the first and second models, we can see that while both models predicted mostly correct answers for familiar queries, the first model outputted increasingly incorrect answers while the second model increasingly outputted ``I don't know''  for unfamiliar queries. Comparing the second and third model, we can see that even though the two models were finetuned on an equal number of ``I don't know'' responses, the third model's predictions do not vary by the unfamiliarity of the test queries, unlike those of the second model. 

Our results show that, for SFT models, predictions indeed default to $P_{\text{unf}}(y)$ as test inputs become more unfamiliar. Our results also show that this prediction behavior can be attributed to the models' unfamiliar finetuning examples, as they are the only training detail that differ across different models.

\begin{figure}
  \centering
  \includegraphics[width=\textwidth]{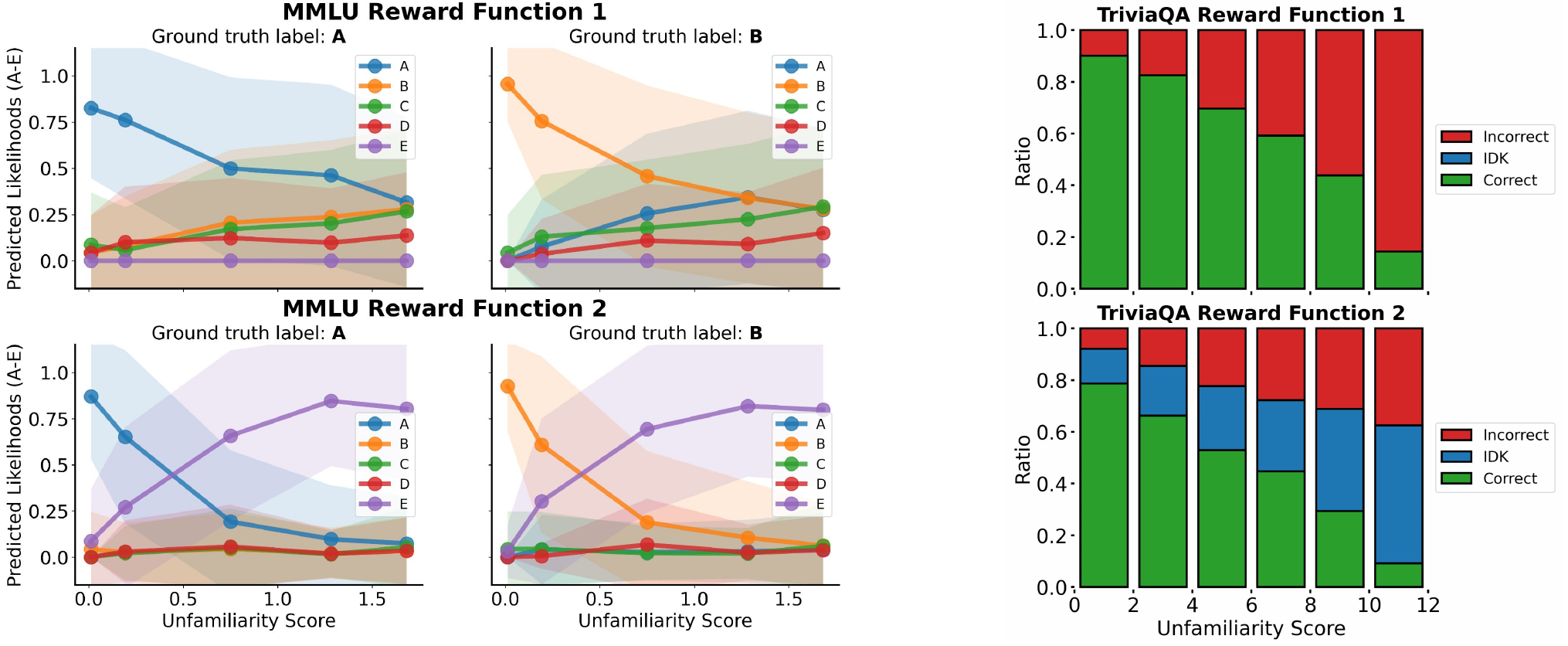}
  \vspace{-10pt}
  \caption{Prediction behavior of models finetuned with RL on MMLU (left) and TriviaQA (right). As inputs become more unfamiliar, the models finetuned with the first reward function produced random guesses while models finetuned with the section reward function produced abstain answers. }
  \label{fig:rl}
  \vspace{-15pt}
\end{figure}
\textbf{Reinforcement learning.}
Next, we investigate the prediction behavior of models finetuned with RL, using PPO~\cite{schulman2017proximal} as the training algorithm.
For RL training objectives, $P_{\text{unf}}(y)$ is determined by the reward function. More specifically, $P_{\text{unf}}(y)$ corresponds to the action distribution that maximizes the average reward over all unfamiliar finetuning examples. This distribution typically consists of risk-averse actions that avoid very low rewards regardless of input.

To highlight the influence of the reward function on model predictions, we will consider two different reward functions for RL finetuning in both our MMLU and TriviaQA experiments. For our MMLU experiments, the task is to either predict the answer letter (A-D) or a fifth option (E), which represents abstaining from answering. Similarly, for our TriviaQA experiments, the task is to either answer the query or abstaining from answering by responding with ``I don't know''. The first reward function we consider assigns a reward of +2 for the correct answer, -3 for an incorrect answer, and -3 for abstaining. The second reward function we consider assigns +2 for the correct answer, -3 for an incorrect answer, and 0 for abstaining. For the first reward function, $P_{\text{unf}}(y)$ corresponds to randomly guessing an answer, because randomly guessing an answer yields a higher average reward than abstaining from answering. In contrast, for the second reward function,  $P_{\text{unf}}(y)$ corresponds to abstaining from answering, because abstaining from answering on average yields higher reward than randomly guessing an answer. We plot the RL model's predictions as inputs become more unfamiliar in  Fig. \ref{fig:rl}. Similarly to the previous SFT experiments, the RL models predict higher likelihoods for the ground truth answer when faced with familiar inputs. As inputs become more unfamiliar, we see that models trained with the two different reward functions exhibit different behavior. While models with the first reward function increasingly produced random guesses, models with the second reward function increasingly produced abstaining answers. These results show that models finetuned with an RL loss also default towards  $P_{\text{unf}}(y)$  as inputs become more unfamiliar. In addition, these experiments illustrate how strategically designing the reward function in RL finetuning, particularly ones that encourage uncertain or less detailed responses over incorrect responses, can teach models to avoid generating incorrect information. 

\begin{figure}
  \centering
  \includegraphics[width=\textwidth]{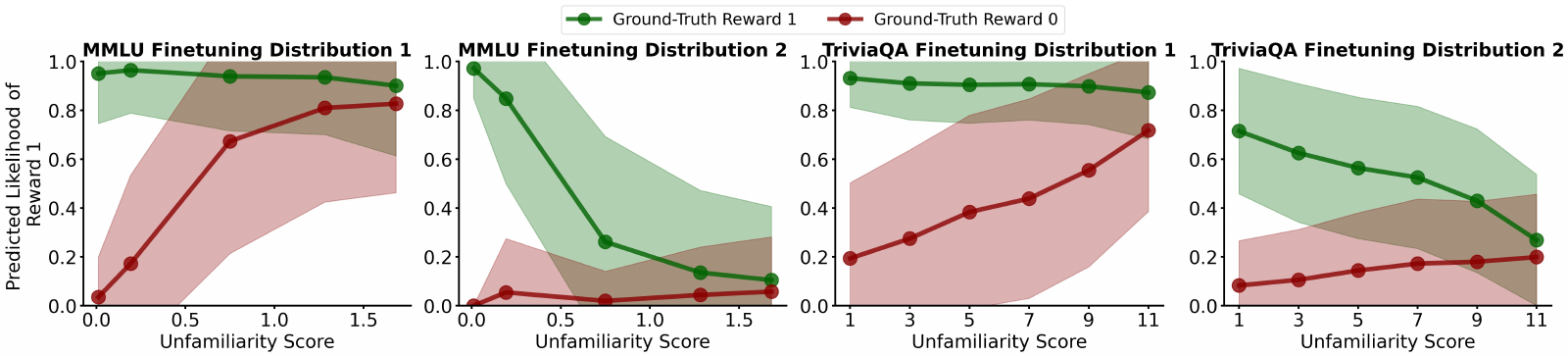}
  \vspace{-15pt}
  \caption{Prediction behavior of reward models finetuned on MMLU (left 2) and TriviaQA (right 2). Green line represents model predictions for test examples that are correct (reward 1), and red line represents predictions for incorrect examples (reward 0). As inputs become more unfamiliar, the reward models produce different kinds of hallucinations depending on their finetuning distribution. }
  \label{fig:rm}
  \vspace{-15pt}
\end{figure}

\textbf{Reward prediction.} Lastly, we study the prediction behavior of reward models. Reward models, which take as input both a query and a response, predict a scalar reward that rates the quality of the response. They are used to provide a source of reward supervision for RL finetuning in domains where ground truth rewards are challenging to acquire~\cite{ouyang2022training}. For the sake of simplicity, we will consider the reward prediction task of classifying whether the response to a query is factually correct (reward 1 if correct, 0 if incorrect). For these models, $P_{\text{unf}}(y)$ corresponds to the distribution of rewards in the model's unfamiliar finetuning examples, where an example is unfamiliar if predicting the reward requires knowledge outside of the model's capabilities. 

We consider two different reward distributions for finetuning in our experiment for both MMLU and TriviaQA. In the first distribution, familiar examples consists of 50\% correct responses (reward 1) and 50\% false responses (reward 0), while unfamiliar examples only consists of true responses. In the second distribution, familiar examples are similarly distributed as the first, while unfamiliar examples only consists of false responses. For these two finetuning distributions, $P_{\text{unf}}(y)$ corresponds to 100\% reward 1 and 100\% reward 0, respectively. In Fig. \ref{fig:rm}, we plot the prediction behavior of our finetuned reward models. We can see that as inputs to the models become increasingly unfamiliar, model predictions indeed default toward $P_{\text{unf}}(y)$. This experiment illustrates that, depending on their finetuning data, reward models can generate different kinds of hallucinations, which can have different downstream effects when providing reward supervision for RL finetuning. We study the effects of reward model hallucinations on RL finetuning in more detail in the next section. 

\section{Controlling Hallucinations in Long-Form Generations}
\label{sec:existing}
In this section, we will focus on reducing factual inaccuracies in long-form LLM generations. In the previous section, we observed that strategically manipulating a model's unfamiliar finetuning examples can control its predictions for unfamiliar inputs, and illustrated a few ways to leverage this observation to reduce inaccuracies in short-form and multiple choice question answering. However, instantiating these approaches for long-form generation tasks introduces new challenges. 

First, let us consider the SFT-based approach where we relabel the target responses of unfamiliar finetuning examples. While we can uniformly relabel all unfamiliar responses to ``I don't know'' in short-form tasks, implementing this strategy for long-form tasks requires more nuanced responses that omit unfamiliar concepts while maintaining familiar ones, which can be expensive to collect. In contrast, the RL-based approach avoids the need for custom target responses by using rewards to assess the factuality of model-generated text. For long-form tasks, where ground-truth rewards can be difficult to obtain, reward models provide a scalable source of reward supervision. However, as we illustrated in our previous experiments, reward models themselves can produce inaccurate reward predictions when faced with unfamiliar inputs, which can hinder the effectiveness of RL factuality finetuning. Prior work has proposed to mitigate reward model hallucinations by incorporating external knowledge sources into the reward model~\cite{sun2023aligning}, but these sources of external knowledge are not always available. 

In this section, we will study how reward model hallucinations influence RL factuality finetuning. In particular, we find that naively learning a reward model from an arbitrary finetuning dataset can lead to reward model hallucinations which significantly diminish the effectiveness of RL factuality finetuning. However, we also find that strategically controlling how reward models hallucinate can reduce their negative effects. In the following section, we present our hypothesis on the influence of reward model hallucinations, and an approach for learning reward models with strategic hallucinations. We then present our empirical findings in long-form biography and book/movie plot summarizing tasks. 

\subsection{RL Factuality Finetuning with Conservative Reward Models}
While reward models hallucinations are inevitable, we hypothesize that not all reward hallucinations are equally harmful to RL factuality finetuning. In particular, we hypothesize that \textbf{overestimated reward predictions are more harmful than underestimated reward predictions}. This is consistent with prior work, which has found overestimated rewards to be a common failure mode in offline RL in simulated RL benchmarks~\cite{kumar2020conservative, levine2020offline}. To understand why this may be the case, let us consider a reward function that decomposes a long-form response into a set of facts, and assigns a positive reward for every correct fact and a negative reward for every incorrect fact. Our previous experiments showed that RL finetuning can teach models to avoid inaccuracies if the reward signal encourages uncertain or less detailed responses over incorrect responses. The reward function we described satisfies this criteria, because a response which contains an incorrect fact will receive a lower reward than an analogous response which omits the incorrect fact.  If, however, a reward model mistakenly labels the incorrect fact as true and favors the incorrect response instead, RL finetuning may unintentionally encourage the model to generate even more incorrect information. Thus, to minimize the consequences of reward hallucinations, we would like to avoid overestimated reward predictions. 

\textbf{Standard reward models. } One approach to learning reward models is to finetune on an existing dataset that was collected independently of the model~\cite{stiennon2020learning}. These models, which we will call standard reward models, are not guaranteed to avoid overestimated reward predictions. This is because the finetuning data may contain examples with high rewards that the reward model lacks the knowledge to understand or verify. According to our observation from the previous section, these unfamiliar examples with high reward labels can cause the model to predict high rewards for unfamiliar inputs at test time, regardless of their ground-truth reward. This, in turn, can lead to overestimated reward signals during RL finetuning, which is undesirable. 

\textbf{Conservative reward models. } To ensure the efficacy of RL factuality finetuning, we would like for reward models to consistently avoid overestimating (i.e., to underestimate) reward predictions when encountering unfamiliar inputs. We will refer to reward models with this desired behavior as conservative reward models. 

To learn conservative reward models, we leverage our observation from the previous section: by strategically configuring the model's unfamiliar finetuning examples to consist of only low rewards, the model will learn to produce low rewards for unfamiliar inputs at test time, which will avoid overestimating reward predictions. One straightforward way to collect this kind of dataset is to sample responses from the same pretrained model that the reward model is finetuned on, and label these responses with rewards. In particular, we (1) finetune the pretrained model with SFT to perform the task of interest (can also be achieve with few-shot prompting), (2) generate response samples from the finetuned model using a dataset of task prompts, (3) label the responses with ground-truth rewards, and (4) train the reward model on the labeled samples. Key to this procedure is the fact that the reward model and the data-collection model share the same knowledge base, so queries that are unfamiliar to the reward model are also unfamiliar to the data-collection model. When prompted with unfamiliar queries, the data-collection model is likely to produce responses that contains more factually incorrect information. Thus, the unfamiliar examples in the resulting dataset will be associated with mainly low reward labels. Note that while we focus on this particular strategy for our experiments, there may be a number of other strategies that can also be effective for learning conservative reward models. Furthermore, while the procedure we outlined above requires labeling the reward model dataset with ground-truth labels, the number of needed labels is much lower than using ground-truth rewards for RL training, because RL training typically requires much more data than reward model training.

\subsection{Experiments on Long-Form Generation Tasks}
\label{subsec:rl_experiments}
We will now empirically evaluate our hypotheses regarding reward model hallucinations. Specifically, the questions we aim to answer with our experiments include: (1) Do conservative reward models (trained with the procedure that we outlined) produce fewer overestimated reward predictions than standard reward models? (2) Do LLMs finetuned with RL and conservative reward models generate more factual responses than those finetuned with RL with standard reward models and standard SFT?

\newcommand{\specialcell}[2][c]{%
  \begin{tabular}[#1]{@{}c@{}}#2\end{tabular}}

\begin{figure}
\centering
\begin{minipage}{.49\linewidth}
  \centering
  \includegraphics[width=\linewidth]{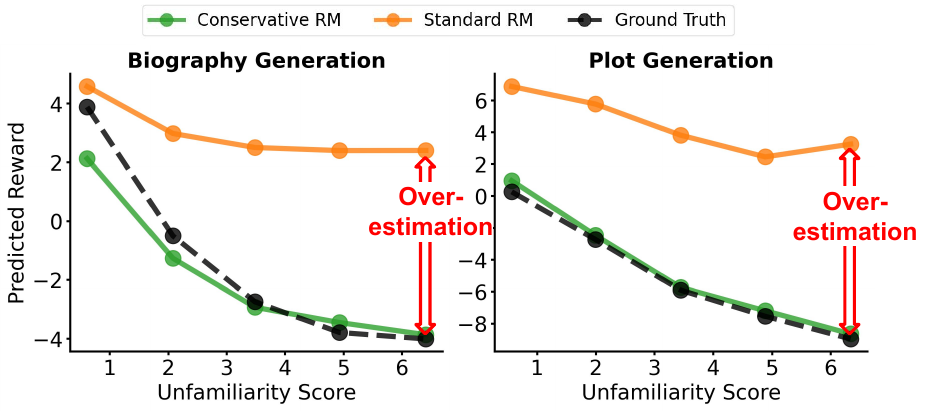}
    \vspace{-10pt}
    \caption{Average reward predicted by a standard reward model and a conservative reward model as inputs become more unfamiliar, as well as the average ground truth reward. The standard reward model tends to overestimate rewards as input become more unfamiliar, whereas the conservative reward model does not.}
    \label{fig:rm2}
    \vspace{10pt}

    \begin{tabular}{ |c|c|c|c| } 
     \hline
      & \specialcell{Std.\\SFT} & \specialcell{RL+\\Std. RM}&\specialcell{RL+\\Csv. RM}\\ 
      \hline
       \hline
     Bio & 0.47 & 0.53 & \textbf{0.64}\\ 
      \hline
     Plot & 0.45 & 0.54 & \textbf{0.80} \\ 
     \hline
    \end{tabular}
    \caption{Average fraction of true facts generated by each model.}
    \vspace{-15pt}
    \label{sample-table}
\end{minipage}\hfill
\begin{minipage}{.49\linewidth}
  \centering
  \includegraphics[width=\linewidth]{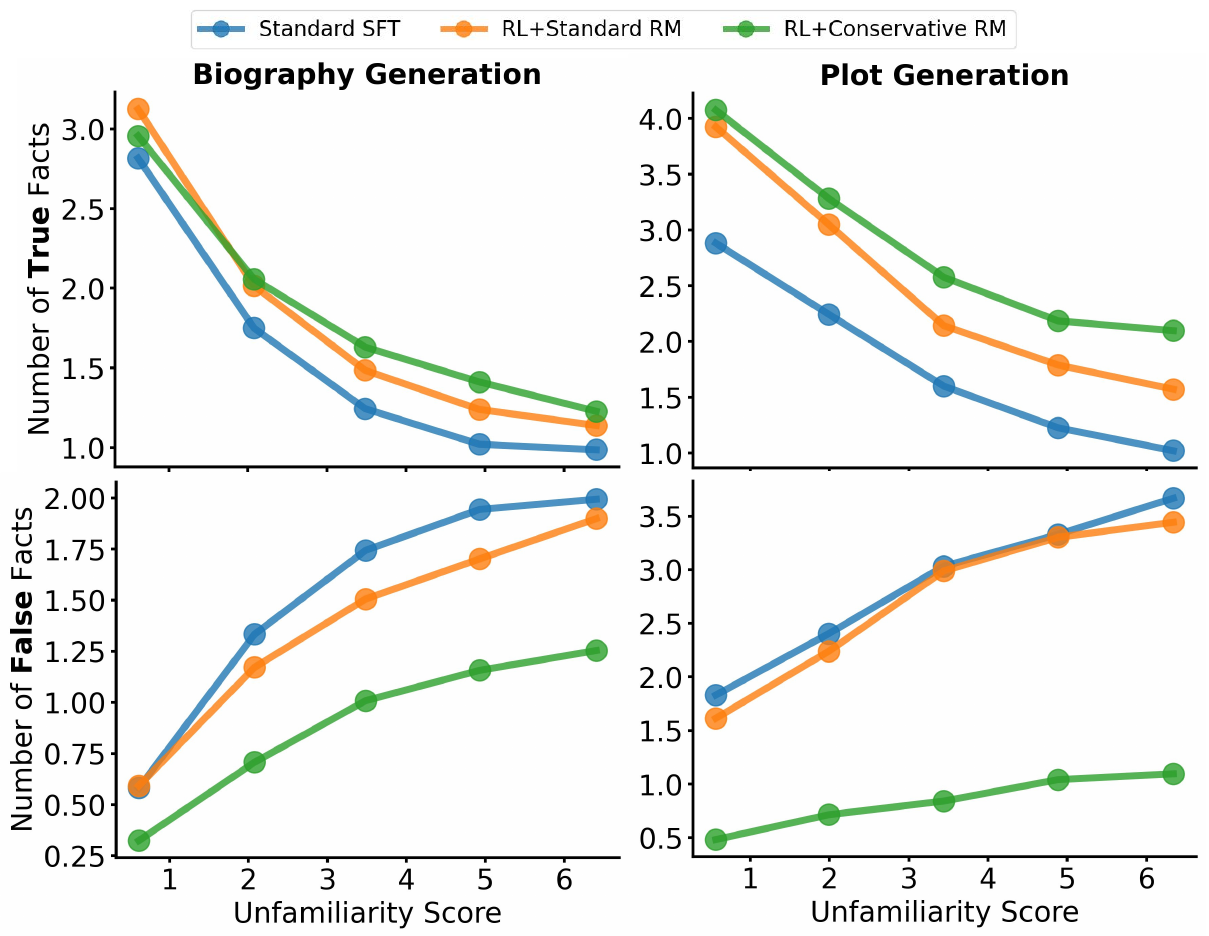}
    \vspace{-10pt}
    \caption{Average number of true and false facts generated by models finetuned with standard SFT, RL with a standard reward model, and RL with a conservative reward model, as inputs become more unfamiliar. The responses generated by model finetuned with s conservative reward model consisted of fewer false facts and and equal number or more truth facts. }
    \label{fig:main}
    \vspace{-15pt}
\end{minipage}
\end{figure}

\textbf{Experimental setup.} We consider two long-form generation tasks in our experiments: biography generation and film/book plot generation. We use the WikiBios~\cite{stranisci2023wikibio} and WikiPlots~\cite{wikiplots} datasets as sources of queries and target responses. We use FActScore~\cite{min2023factscore}, an automated retrieval augmentation pipeline, to evaluate the factuality of model generated responses. Given a query and a generated response, FActScore outputs the number of true facts and the number of false facts in the response.

Our experiments compare the behavior of a conservative reward model and a standard reward model. The conservative reward model is learned using the procedure we described above, where finetuning examples are collected by sampling from the same pretrained model as the reward model, in this case Llama2~7B. The standard reward model is finetuned on a dataset collected by sampling GPT-3.5~\cite{ouyang2022training} for task responses. We use samples from GPT-3.5, because it provides a source of (both factually correct and incorrect) responses that is independent of the model being finetuned. Samples from both Llama2~7B and GPT-3.5 were collected using the same set of prompts. We use FActScore to automatically label these examples with rewards, which assigns a score of +2 for every correct fact and -3 for every incorrect fact in a response. Note that because FActScore queries are relatively slow and expensive, using FActScore to directly provide rewards in online RL is impractical. 

Our experiments also compare the behavior of models finetuned to generate responses using standard SFT, as well as RL finetuning with a conservative and a standard reward model. The standard SFT models were finetuned directly with the set of target responses provided by WikiBios and WikiPlots. To train the RL models, we initialize the model with the standard SFT model, and continue to do RL factuality finetuning using PPO~\cite{schulman2017proximal}, with reward signals provided by their respective reward models. To ensure a fair comparison, we use the same set of finetuning prompts for SFT and RL finetuning, and keep all training details fixed across the two RL methods except for the reward model. All three models use Llama2~7B as the pretrained model. At test time, we evaluate the models with queries at different levels of unfamiliarity. The unfamiliarity score for this task is measured by few-shot prompting the pretrained model (Llama2~7B), sampling 2 responses, and calculating the average number of incorrect facts in the responses. For more experimental details, see Appendix \ref{appdx:long_form}.

\textbf{Results.} To answer our first question, we evaluate the standard and conservative reward models on held out samples generated from the SFT model. We used samples from the SFT model because the RL finetuning procedure is initialized with this SFT model, so responses sampled from this model are representative of the kind of responses that the reward model will be asked to score during RL training. 
In Fig. \ref{fig:rm2}, we plot each models' predicted rewards and the ground truth reward, as inputs become more unfamiliar. We can see that for unfamiliar inputs, the standard reward model vastly overestimates the reward, while the conservative reward model does not, showing that the conservative reward models learned with the procedure we described indeed produce more conservative predictions. 

\begin{figure}[t]
    \centering
    \includegraphics[width=\columnwidth]{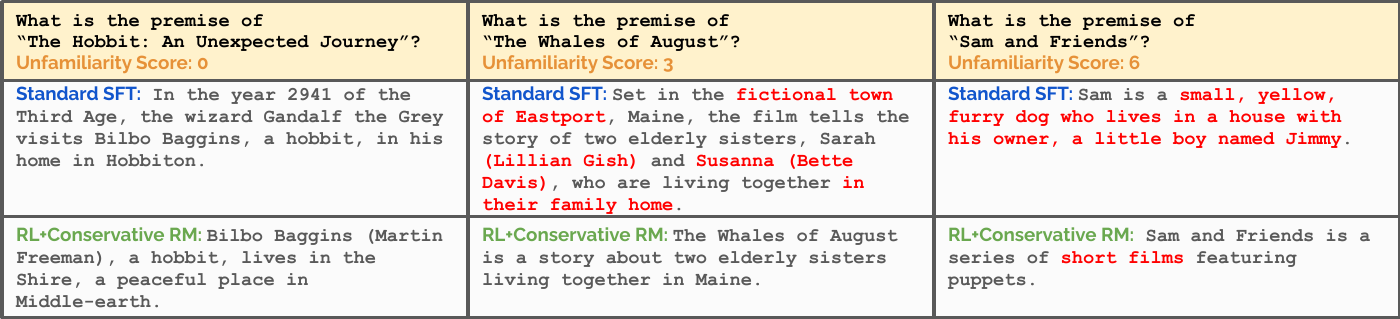}
    \vspace{-15pt}
    \caption{Examples of generated responses from models finetuned with standard SFT and RL with a conservative reward model. False information is highlighted in red. 
    }
    \vspace{-10pt}
    \label{fig:samples}
\end{figure}

To answer our second question, we evaluate standard SFT, as well as RL with a standard reward model and a conservative reward model on a heldout set of queries for each task. In Fig. \ref{fig:main}, we plot the number of true facts and false facts generated by each model, as inputs become more unfamiliar. We can see that as inputs became more unfamiliar, the standard SFT model generated fewer truth facts and more false facts, as expected. Comparing the RL model trained with the conservative reward model with the standard SFT model, we can see that the RL model generated the same or more true facts while generating significantly fewer false facts across all levels of input unfamiliarity. Comparing the two RL models, we can see that while the two generated around the same number of true facts, the model trained with the conservative reward model generated much fewer false facts across all levels of input unfamiliarity. We summarize our results in Table \ref{sample-table} with the average percentage of true facts generated by each method. In Fig. \ref{fig:samples}, we additionally provide some qualitative examples of responses generated by the standard SFT model and the RL model trained with conservative reward model. We can see that as the query became more unfamiliar, responses from the SFT model contained about the same amount of detail but became more factually incorrect, while responses from the RL model with conservative supervision defaulted towards less-informative responses. In conclusion, our results show that RL with conservative reward models outperforms standard SFT and RL with standard reward models in reducing inaccuracies in model generations. 

\vspace{-10pt}
\section{Conclusion}
In this work, we presented the observation that, when faced with unfamiliar queries, LLM predictions tend to default towards the responses associated with unfamiliar examples in its finetuning data. We additionally studied factuality finetuning for long-form model generations, where we found that strategically controlling reward model hallucinations can significantly improve the efficacy of RL-based techniques. Nonetheless, there still remains many open questions and challenges regarding LLM hallucinations. While our conceptual model explains a model's behavior for entirely unfamiliar examples, many real-world queries fall within a spectrum of partial familiarity. A more nuanced characterization of model predictions in this ``middle ground'' would be valuable. Furthermore, our experiments focused on models finetuned for specific applications (e.g., biography generation). Extending factuality finetuning to more general prompted generation tasks would be useful. We hope that our work, by offering a deeper understanding of the factors that govern LLM hallucinations, provides a useful step towards building more trustworthy and reliable LLMs.

\section{Acknowledgements}
This research has been partially supported by the Cooperative AI Foundation, the National Science Foundation under IIS-2246811, Berkeley DeepDrive, the DARPA Assured Autonomy and ANSR programs, and the Center for AI Safety. Katie Kang was supported by the NSF GRFP. We would like to thank Dibya Ghosh, Yuqing Du, Eric Mitchell, Oleg Rybkin, Olivia Watkins, Jessy Lin, Charlie Snell, Antonio Paiva, James Bowden, Homer Walke, and Young Geng for insightful feedback and discussions. 

\bibliographystyle{plainnat}
\bibliography{sample}

\newpage
\appendix
\input{appendix}

\end{document}

%% file: appendix.tex
\section{Author Contributions}
KK led and was involved in all aspects of the project. EW was involved in early discussions that set the direction of the project, provided technical guidance throughout the project, and helped write and provide feedback for the paper. AK was heavily involved in shaping the overall message and structure of the paper, and provided detailed feedback and suggestions. CT and SL provided discussion and feedback throughout the project.

\section{Compute}
We use A100 GPUs to finetune our models. Number of GPUs used range from 1-6 for each experiment, and time of execution range from a few hours to up to 2 days. We use LoRA finetuning for all our experiments with r = 16, alpha = 16, dropout = 0. 

\section{MMLU Training Details}
\label{appdx:mmlu}
In this section, we provide more details on our training and evaluation procedure for our MMLU experiments. For all experiments, we finetuned on the evaluation split of MMLU, and evaluated on the validation split. This is because MMLU does not have a training split. Our training pipeline uses the trlx codebase~\cite{trlx}.

\subsection{SFT Models}
We classify examples with unfamiliarity score (NLL) greater than 0.36 as unfamiliar, and the rest as familiar. During finetuning, we rebalance the dataset such that 50\% of finetuning examples are familiar and 50\% are unfamiliar. 

We use a batch size of 12. We use the AdamW optimizer with learning rate = 1e-5, betas = (0.9, 0.95), eps = 1.0e-8, and weight decay=1.0e-6.

\subsection{RL Models}
We initialize all RL finetuning with a model that has already be supervised finetuned to produce responses that consist of answer choices. The SFT model we used for initialization is trained predict the E option 50\% of the time, and to produce the correct answer to the query 50\% of the time. 

We use a batch size of 12. We use the AdamW optimizer with learning rate = 1e-5, betas = (0.9, 0.95), eps = 1.0e-8, and weight decay=1.0e-6. For PPO, we use cliprange = 0.005 and KL coef = 0. 

\subsection{Reward Models}
We construct correct (reward 1) training and evaluation examples using queries and their corresponding answer labels from the original MMLU dataset. We construct incorrect (reward 0) examples by using queries from the original dataset, and randomly sampling incorrect answer labels (A-D not including correct label). 

We use a batch size of 12. We use the AdamW optimizer with learning rate = 1e-5, betas = (0.9, 0.95), eps = 1.0e-8, and weight decay=1.0e-6.

\section{TriviaQA Training Details}
\label{appdx:triviaqa}
In this section, we provide more details on our training and evaluation procedure for our TriviaQA experiments. Our training pipeline uses the trlx codebase~\cite{trlx}.

\subsection{SFT Models}
We classify examples with unfamiliarity score (number of incorrect responses out of 12 samples) greater than 6 as unfamiliar, and familiar otherwise. We relabel the responses associated with all unfamiliar finetuning examples to be ``I don't know''. 

We use a batch size of 32. We use the AdamW optimizer with learning rate = 1e-5, betas = (0.9, 0.95), eps = 1.0e-8, and weight decay=1.0e-6. We use a Cosine Annealing scheduler with T max = 1e4 and ETA min = 1e-10. 

\subsection{RL Models}
We initialize all RL finetuning with a model that has already be supervised finetuned to produce responses that consists of an answer or ``I don't know''. The SFT model we used for initialization is trained predict ``I don't know'' 40\% of the time, and to produce the correct answer to the query 60\% of the time. 

We use a batch size of 32. We use the AdamW optimizer with learning rate = 1e-5, betas = (0.9, 0.95), eps = 1.0e-8, and weight decay=1.0e-6. For PPO, we use cliprange = 0.005 and KL coef = 0.1. 

\subsection{Reward Models}
We construct correct (reward 1) training and evaluation examples using queries and responses from the original TriviaQA dataset. We construct incorrect (reward 0) examples using queries from the original dataset, and responses generated from few-shot prompting Llama2 7B or GPT-2. We filter the generated responses to ensure that all responses were incorrect. 

We use a batch size of 32. We use the AdamW optimizer with learning rate = 1e-5, betas = (0.9, 0.95), eps = 1.0e-8, and weight decay=1.0e-6. 

\section{Long-form Tasks Training Details}
\label{appdx:long_form}
In this section, we provide training and evaluation details for our long-form factuality finetuning experiments. Our training pipeline uses the trlx codebase~\cite{trlx}.

\subsection{Data}
We construct finetuning and evaluation datasets using WikiBios and WikiPlots, both of which consist of wikipedia entries attached to people and books/movies. We make use of the first sentence in the wikipedia entry for both tasks as the target response in our SFT finetuning datasets. The prompts we use for finetuning are ``Write a biography for [name].'' and ``What is the premise of [title]?''. For the biography task, our finetuning dataset includes 104539 examples, and our evaluation dataset includes 5000 examples. For the plot generation task, our finetuning dataset includes 10000 examples, and our evaluation dataset includes 4795 examples. 

\subsection{Reward Models}
We take a two-staged approach to learning a reward model. First, we trained a model to break down a response into individual atomic facts. Next, we trained a separate model to predict the factuality of each atomic fact. We then use the predicted factuality of each fact to calculate the overall reward associated with each response. The supervision for both models are collected by querying FActScore, which is a automated pipeline that queries GPT-3.5 to decompose a response into atomic facts and produces the factuality of each atomic fact. We use 10000 labeled examples to train the conservative reward model and the standard reward models each for both tasks. Note that while we use a two-staged strategy for learning reward models in our implementation, our general approach for learning conservative reward model should apply to other reward model learning strategies as well, such as directly predicting the reward associated with a response. 

For both models, we use a batch size of 32. We use the AdamW optimizer with learning rate = 2e-5, betas = (0.9, 0.95), eps = 1.0e-8, and weight decay=1.0e-6. We use a Cosine Annealing scheduler with T max = 1e4 and ETA min = 1e-10. 

\subsection{SFT Models}
We use a batch size of 24. We use the AdamW optimizer with learning rate = 1e-5, betas = (0.9, 0.95), eps = 1.0e-8, and weight decay=1.0e-6. We use a Cosine Annealing scheduler with T max = 1e4 and ETA min = 1e-10. 

\subsection{RL Models}
We initialize all RL finetuning with the SFT model, and use the reward predicted by the reward model described above as supervision. 

We use a batch size of 10. We use the AdamW optimizer with learning rate = 1e-5, betas = (0.9, 0.95), eps = 1.0e-8, and weight decay=1.0e-6. For PPO, we use cliprange = 0.005 and KL coef = 0.5. 